\documentclass{article}
\usepackage{spconf,amsmath,graphicx}
\usepackage{hyperref}       
\usepackage{url}            
\usepackage[dvipsnames]{xcolor}
\usepackage{booktabs}

\usepackage{amsmath}
\usepackage{amssymb}
\usepackage{mathtools}
\usepackage{amsthm}
\usepackage{multirow}
\usepackage{newtxtext,newtxmath}
\usepackage[symbol]{footmisc}

\DeclareMathOperator{\R}{\mathbb{R}}
\DeclareMathOperator{\E}{\mathcal{E}}
\DeclareMathOperator{\D}{\mathcal{D}}

\newcommand{\etal}{\textit{et al.~}}

\newcommand\blfootnote[1]{%
  \begingroup
  \renewcommand\thefootnote{}\footnote{#1}%
  \addtocounter{footnote}{-1}%
  \endgroup
}

\name{\begin{tabular}{c} 
Eliya~Nachmani$^{\,\star}$ \qquad Alon~Levkovitch$^{\,\star, \diamond}$ \qquad Yifan~Ding$^{\,\dagger}$ \qquad Chulayuth~Asawaroengchai$^{\,\star}$  \\[1mm]
\qquad Heiga~Zen$^{\,\dagger}$ \qquad Michelle~Tadmor~Ramanovich$^{\,\star}$
\end{tabular}
}
\address{$^{\star}$ Google Research, Israel \quad 
  $^{\dagger}$Google DeepMind, Japan
 \quad 
  $^{\diamond}$Tel-Aviv University
\\[1mm]
\texttt{\small 
\{eliyn,~alevkovitch,~dyf\}@google.com
}
}

\title{Translatotron 3: Speech to Speech Translation with Monolingual Data}

\begin{document}
\ninept
\maketitle
\begin{abstract}
This paper presents \emph{Translatotron~3}, a novel approach to unsupervised direct speech-to-speech translation from monolingual speech-text datasets by combining masked autoencoder, unsupervised embedding mapping, and back-translation. Experimental results in speech-to-speech translation tasks between Spanish and English show that Translatotron~3 outperforms a baseline cascade system, reporting $18.14$ BLEU points improvement on the synthesized Unpaired-Conversational dataset. In contrast to supervised approaches that necessitate real paired data, or specialized modeling to replicate para-/non-linguistic information such as pauses, speaking rates, and speaker identity, Translatotron~3 showcases its capability to retain it.
\end{abstract}


\blfootnote{$\diamond$ Work done during internship at Google.}

\section{Introduction}
\vspace{-2mm}
\label{sec:intro}

Recent years have seen significant advancements in direct Speech-To-Speech Translation (S2ST) \cite{jia2022translatotron}. Major motivations of direct S2ST is the potential to preserve and translate para-/non-linguistic speech characteristics, such as speaking styles, emotions, emphasis, phonation, and vocal bursts. Although these audible characteristics are essential aspects of human verbal communication, they are often lost in traditional cascade speech translation systems \cite{lavie1997janus,wahlster2013verbmobil,nakamura2006atr}. 
Previous S2ST research, such as \cite{jia2019direct, jia2022translatotron, lee2021direct, lee-etal-2022-direct}, primarily used supervised learning that rely on bilingual speech datasets. This dependency introduces two limitations; (1) Supporting low-resource languages is difficult as collecting bilingual speech datasets including these languages is hard, and (2) due to the lack of bilingual speech datasets with corresponding para-/non-linguistic characteristics in both source and target languages, para-/non-linguistic characteristics in the source speech cannot be transferred to the translated speech. This paper addresses the problem of unsupervised S2ST, which can eliminate the requirement for bilingual speech datasets. Previous research in Unsupervised Machine Translation (UMT) has employed techniques such as back-translation \cite{sennrich2015improving}, where a synthetic translation of the source language is used as ``bilingual text datasets'' \cite{artetxe2018unsupervised,lample2018unsupervised}. The proposed approach, called \emph{Translatotron~3}, incorporates (i) pre-training the entire model as a masked autoencoder \cite{he2022masked} with SpecAugment \cite{park2019specaugment}, (ii) unsupervised embedding mapping based on the multilingual unsupervised embeddings (MUSE) \cite{conneau2017word}, and (iii) a reconstruction loss based on back-translation \cite{sennrich2015improving}, to train an encoder-decoder direct S2ST model from Translatotron~2 \cite{jia2022translatotron} in a fully unsupervised manner. 
The model has a shared encoder and two decoders, one for each language. Training consists of two phases. The first phase trains the model as a masked autoencoder \cite{he2022masked} using reconstruction loss with the unsupervised MUSE loss \cite{lample2017unsupervised}. In the second phase the encoded features are transferred into the target language and back via back-translation to compute a reconstruction loss.
The main contributions of this work are as follows: (i) The first fully unsupervised end-to-end model for direct speech to speech translation trained on real speech dataset. (ii) Outperforming a cascade baseline for unsupervised S2ST by large margin in two synthesized and real speech datasets and approaches supervised models for English to Spanish translation on the CVSS dataset. (iii) Demonstrating the ability to transfer para-/non-linguistic characteristics such as pauses, speaking rates and speaker identity, from the source speech through experimental on a real speech. 
\vspace{-6mm}
\section{Related works} 
\vspace{-2mm}
\label{sec:related_works}
Conventional S2ST systems such as \cite{lavie1997janus, wahlster2013verbmobil,nakamura2006atr} were implemented as a cascade of three separate components: Automatic Speech Recognition (ASR) for the source language, Machine Translation (MT) for converting source language text to target language text, and Text-To-Speech (TTS) synthesis for generating speech from the translated text. Jia \etal \cite{jia2019direct} proposed the first end-to-end direct S2ST model called \emph{Translatotron}. Subsequently, \emph{Translatotron~2} \cite{jia2022translatotron} was proposed as an improvement on the original Translatotron model, offering better performance, controllability, and robustness. Moreover, Zhang \etal \cite{zhang2021uwspeech}, Lee \etal \cite{lee2021textless}, Huang \etal \cite{huang2022transpeech} used different forms of discrete representations for S2ST. There are other approaches based on self-supervision to use untranscribed speech and unspoken text datasets. Tang \etal \cite{tang2022unified} used two sub-tasks for pre-training, one for untranscribed speech data and another for unspoken text data. Dong \etal \cite{dong2022leveraging} proposed a system which is based on a cascaded system which generates pseudo-labels. There are also approaches based on self-supervised learning techniques to leverage untranscribed speech datasets. Kano \etal \cite{kano2021transformer} introduced an S2ST model with a cascade of three autoregressive decoders. Wang \etal \cite{wang2022simple} proposed an approach that combines teacher models and pseudo-labeling to utilize unlabelled data. Inaguma \etal \cite{inaguma2022unity} proposed UnitY, a S2ST model using a two-pass architecture. 
Wei \etal \cite{wei2022joint} proposed an S2ST model that is jointly pre-trained with monolingual speech and bilingual text datasets. Another line of research involves using speech tokens \cite{zeghidour22-soundstream,chung2021w2v,defossez2022high,baevski2020wav2vec,hsu2021hubert, tjandra2019speech} with Language Models (LM) for supervised speech-to-speech translation \cite{rubenstein2023audiopalm,li2023textless,inaguma2022unity,kim2023many,wei2023joint} and speech-to-text translation \cite{zhang2023dub,ao2021speecht5,zhang2022yitrans,zhang2022speechlm}. Unsupervised machine translation (UMT) aims to perform MT without the use of bilingual text datasets. Recent work \cite{artetxe2018unsupervised,lample2018unsupervised} has shown promising results UMT. Both approaches were built upon unsupervised cross-lingual embedding mappings \cite{artetxe2017learning, artetxe2018robust,conneau2017word}. Furthermore, \cite{chung2019towards} train unsupervised speech to text translation models based on unsupervised dictionary.

\section{Translatotron~3}
\vspace{-2mm}
\label{sec:model}

\begin{figure*}[t]
\centering
\begin{tabular}{c@{~}c}
\includegraphics[width=.5\textwidth, height=.3\textwidth]{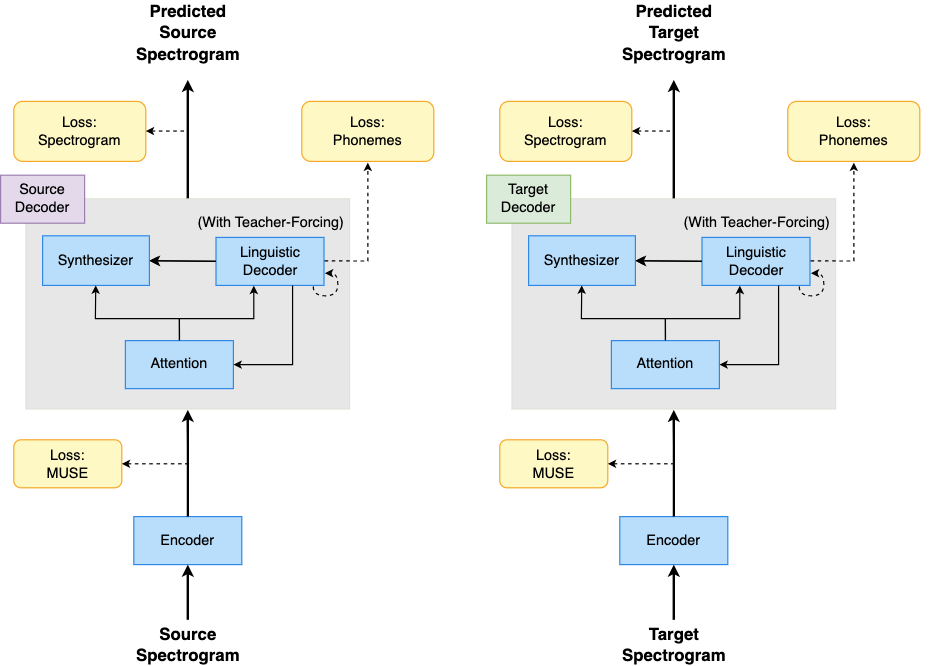} &
\includegraphics[width=.5\textwidth, height=.3\textwidth]{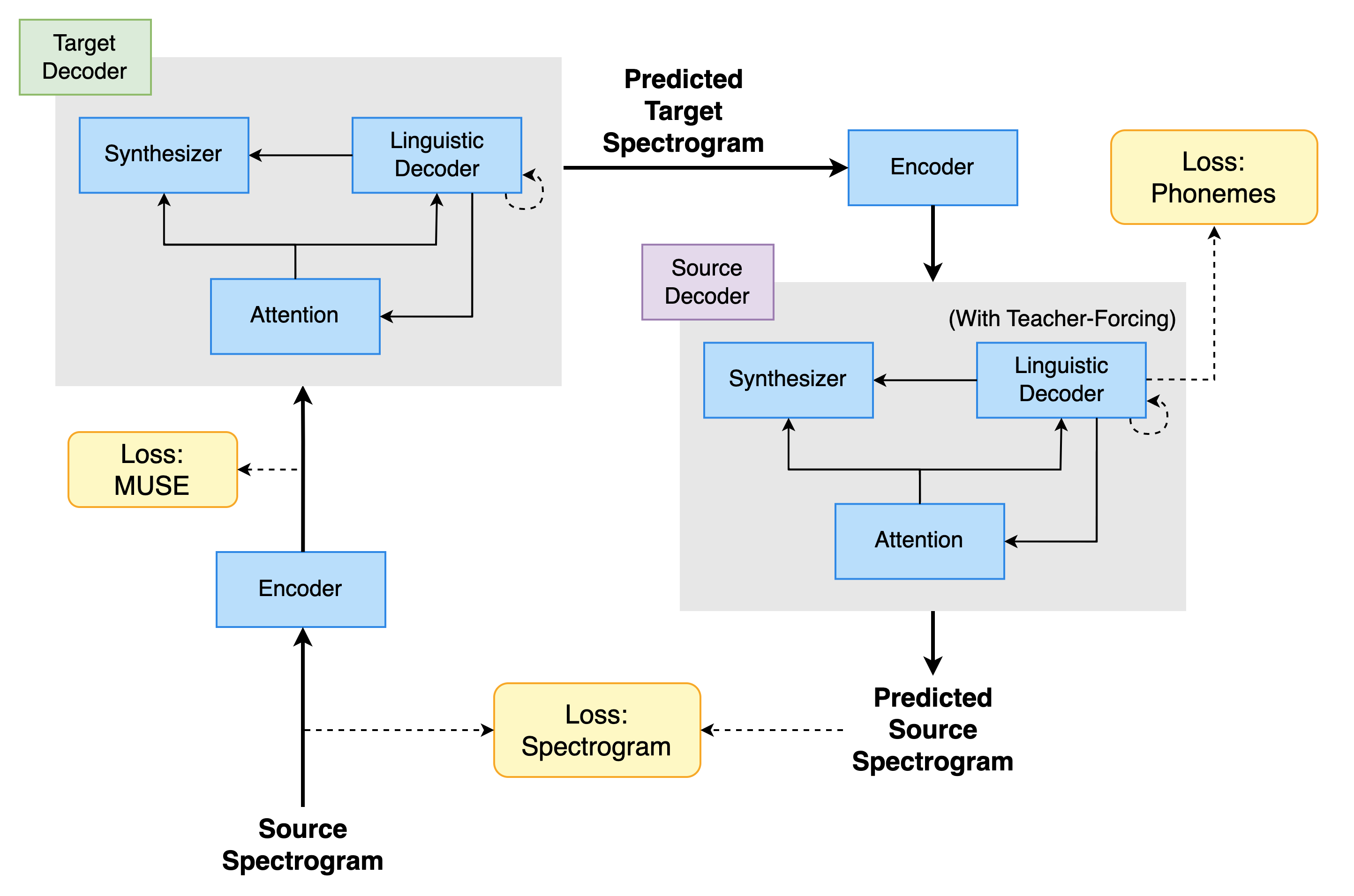}\\
(i) & (ii)\\
\end{tabular}
  \caption{The two training phases in the proposed approach. (i) Phase 1 uses the reconstruction loss via the auto-encoding path. (ii) Phase 2 employs the reconstruction loss via back-translation}
\label{fig:arch2}
\vspace{-3mm}
\end{figure*}

The proposed approach, Translatotron~3, adopts a novel architecture to allow unsupervised S2ST where there are a shared encoder and two decoders for the source and target languages. The model is trained using a combination of the unsupervised MUSE embedding loss, reconstruction loss, and S2S back-translation loss. During inference, the shared encoder encode the input into a multilingual embedding space, which is subsequently decoded by the target decoder. 
\subsection{Architecture} Translatotron~3 employs a shared encoder $\mathcal{E}$ to encode both the source and target languages. The decoder $\mathcal{D}$ is composed of a linguistic decoder, an acoustic synthesizer, and a singular attention module. There are two decoders, one for the source language $\mathcal{D}_s$ and another for the target language $\mathcal{D}_t$. 
\textbf{Encoder}: The encoder $\mathcal{E}$ has the same architecture as the speech encoder in Translatotron~2 \cite{jia2022translatotron}. The output of the encoder $\mathcal{E}(S^{in})$ is split into two parts $\mathcal{E}(S^{in}) = \left [\mathcal{E}_{m}(S^{in}), \mathcal{E}_{o}(S^{in})\right ],$ where $S^{in}$ can be the source or target language. The first half of the output $\mathcal{E}_{m}(S^{in})$ is trained to be the MUSE embeddings of the text of the input spectrogram $S^{in}$. This is forced using the MUSE loss. The latter half $\mathcal{E}_{o}(S^{in})$ is updated without the MUSE loss. It is important to note that the same encoder $\mathcal{E}$ is shared between source and target languages. Furthermore, the MUSE embeddings are multilingual in nature. As a result, the encoder is able to learn a multilingual embedding space across source and target languages. 
\textbf{Decoder}: The decoder $\mathcal{D}$ is composed of three distinct components, namely the linguistic decoder, the acoustic synthesizer, and the attention module. To effectively handle the different properties of the source and target languages, it has two separate decoders, $\mathcal{D}^s$ and $\mathcal{D}^t$, for the source and target languages, respectively. The decoder output can be formulate as ${S^{out}} = \mathcal{D}^{out}\left(\mathcal{E}(S^{in})\right),$ where $S^{in}$ and $S^{out}$ correspond to the input and output sequences. Both $S^{in}$ and $S^{out}$ may represent either the source or target language, as well as $\mathcal{D}^{out}$.
\subsection{Training Objective}
Fig.~\ref{fig:arch2} illustrates the training methodology of the proposed approach. It consists of two phases: (i) auto-encoding, reconstruction phase and (ii) back-translation phase. In the first phase, the network is trained to auto-encode the input to a multilingual embedding space using the MUSE loss and the reconstruction loss. This phase aims to ensure that the network generates meaningful multi-lingual representations. In the second phase, the network is further trained to translate the input spectrogram by utilizing the back-translation loss. To mitigate the issue of catastrophic forgetting and enforcing the latent space to be multilingual, the MUSE loss and the reconstruction loss are also applied in the second phase of the training. To ensure that the encoder learns meaningful properties of the input, rather than simply reconstructing the input, we apply SpecAugment \cite{park2019specaugment}. The first  phase can be viewed as masked auto-encoder training \cite{he2022masked}.
\subsubsection{MUSE Loss} To ensure that the encoder $\E$ generates multi-lingual representations that are meaningful for both decoders $\D^s$ and $\D^t$, we employ a MUSE loss during training. The MUSE loss forces the encoder to generate such a representation by using pre-trained MUSE embeddings \cite{conneau2017word}. These embeddings are computed in an unsupervised manner. During the training process, given an input transcript with $n$ words, we extract $n$ corresponding MUSE embeddings $E \in \R^{n\times d}$ from the embeddings of the input language. The error between $E$ and the $n$ first output vectors of the encoder $\E$ is then minimized. This can be represented as $\mathcal{L}_{\text{MUSE}}(S^{in}) = \frac{1}{n}\sum_{i=1}^n \left\|\E(S^{in})_i - E_i\right\|_2^2,$

where $S^{in}$ represents the input spectrogram, which may be in the source or target language, $E_i$ is the $d$-dimensional embedding vector for $i$-th word. Note that the encoder is indifferent to the language of the input during inference due to the multilingual embeddings. 
\subsubsection{Reconstruction Loss}
Fig.~\ref{fig:arch2}(a) illustrates the reconstruction training phase.
In this phase, the network learns to auto-encode both the source and target languages. 
The reconstruction loss is computed as a linear combination of three losses for both the source and target languages. The main reconstruction objective is $\mathcal{L}_{spec}$, where given a paired example $\{{S^t}' , S^t \}$, the loss function can be expressed as 
\begin{equation}
    \mathcal{L}_\text{spec} \left ( {S^{t}}', {S^{t}} \right )  = \frac{1}{TK} \sum_{i=1}^T \sum_{j=1}^2 \left \|{S^{t}_i}' - S^t_i\right \|_j^j,
\end{equation}

where $S^{t}_i$ denotes the $i$-th frame of $S^t$, $T$ is the number of frames in $S^t$, $K$ is the number of frequency bins in $S^{t}_i$, and $\left\| \cdot \right\|_j^j$ denotes the $L_j$ distance. The second loss is a duration loss between total number of frames $T$ and the sum of phoneme durations predicted in the acoustic synthesizer. It is given as $\mathcal{L}_\text{dur} = \left( T - \sum_{i=1}^p d_i \right)^2,$ where $d_i$ is the predicted duration for the $i$-th phoneme. The last loss is the auxiliary phoneme loss. Let $\tilde{P}^t = \{ \tilde{P}^t_1, \dots, \tilde{P}^t_p\}$ be the sequence of predicted probabilities over target phonemes, $P^t = \{ P^t_1, \dots, P^t_p \}$ be the ground-truth target phoneme sequence, and $\text{CE}(\cdot,\cdot)$ be the cross entropy. Then we have 
\begin{equation}
    \mathcal{L}_\text{phn} \left (\tilde{P}^t, P^t \right ) = \frac{1}{P} \sum_{i=1}^{p} \text{CE}\left (\tilde{P}^t_i, P^t_i\right ).
\end{equation} The overall reconstruction loss can be summarized
\begin{align}
    \mathcal{L}_{\text{recon}} &= \mathcal{L}^{src}_{\text{spec}}\left ( {S^{s}}', {S^{s}} \right ) + \mathcal{L}^{src}_{\text{dur}} + \mathcal{L}^{src}_{\text{phn}}\left ( \tilde{P}^s, P^s \right ) \notag \\ &+
    \mathcal{L}^{tgt}_{\text{spec}}\left ( {S^{t}}', {S^{t}} \right ) + \mathcal{L}^{tgt}_{\text{dur}} + \mathcal{L}^{tgt}_{\text{phn}}\left ( \tilde{P}^t, P^t \right ),
    \label{eqn:recon_loss}
\end{align}
where $S^{s}$ and $S^{t}$ are the source and target spectrograms respectively, $S^{s'}$ and $S^{t'}$ represents the model spectrogram output predictions, $\tilde{P}^s$ and $\tilde{P}^t$ represent the probability distributions of the phonemes of the source and target languages respectively, and  $P^s$ and $P^t$ represent the phoneme sequences of the source and target text respectively.
\subsubsection{Back-Translation Loss} Fig.~\ref{fig:arch2}(b) illustrates the back-translation training phase. The process begins with the encoding of the source input spectrogram, represented as $\mathcal{E}(S^{s})$, followed by the use of the target language decoder to produce a pseudo-translation, denoted as $\hat{S}^{t'}$ as $\hat{S}^{t'} = \mathcal{D}^t\left(\mathcal{E}(S^{s})\right).$
Subsequently, we proceed to encode the pseudo-translation, utilizing the encoder $\mathcal{E}(\hat{S}^{t'})$. The final step in this process involves decoding the encoded pseudo-translation using the decoder of the source language $\hat{S}^{s} = \mathcal{D}^s\left(\mathcal{E}(\hat{S}^{t'})\right).$
Lastly, we apply a loss function to minimize the dissimilarity to the input spectrogram
\begin{equation}
    \mathcal{L}^{src2tgt}_{\text{back-translation}} =  \mathcal{L}^{src}_{\text{spec}}\left ( \hat{S}^{s}, {S^{s}} \right ) + \mathcal{L}^{src}_{\text{dur}} + \mathcal{L}^{src}_{\text{phn}}\left ( \tilde{P}^s, P^s \right ).
\end{equation}
We also apply the aforementioned process in the reverse direction, from the target to the source language. This process entails utilizing the same methodologies and techniques as previously described, with the only difference being the direction of the translation
\begin{equation}
    \mathcal{L}^{tgt2src}_{\text{back-translation}} = \mathcal{L}^{tgt}_{\text{spec}}\left ( \hat{S}^{t}, {S^{t}} \right ) + \mathcal{L}^{tgt}_{\text{dur}} + \mathcal{L}^{tgt}_{\text{phn}}\left ( \tilde{P}^t, P^t \right ).
\end{equation}
The overall back-translation loss is given as
\begin{equation}
    \label{eq:back_translation}
    \mathcal{L}_{\text{back-translation}} = \mathcal{L}^{src2tgt}_{\text{back-translation}} + \mathcal{L}^{tgt2src}_{\text{back-translation}}.
\end{equation}
Note that $\mathcal{L}^{src}_{\text{spec}} ( \hat{S}^{s}, {S^{s}} ) + \mathcal{L}^{src}_{\text{spec}} ( \hat{S}^{t}, {S^{t}} )$, which is a part of \\ $\mathcal{L}_{\text{back-translation}}$, can be viewed as a cycle consistency loss proposed for unpaired image-to-image translation \cite{zhu2017unpaired}.
\subsubsection{Overall Loss} For the reconstruction phase, the loss is given by
\begin{equation}
    \mathcal{L}_{\text{recon-phase}} = \mathcal{L}_{\text{recon}} + \mathcal{L}_{\text{MUSE}}(S^{s}) + \mathcal{L}_{\text{MUSE}}(S^{t})
\end{equation}
In the back-translation phase, the optimization of both the back-translation loss ($\mathcal{L}_{\text{back-translation}}$), and the reconstruction loss ($\mathcal{L}_{\text{recon-phase}}$), is carried out to ensure that the encoder output produces a multilingual embedding. The overall loss is:
\begin{equation}
    \mathcal{L}_{\text{BT-phase}} = \mathcal{L}_{\text{back-translation}} + \mathcal{L}_{\text{recon-phase}}.
\end{equation}

\section{Experiments and Results}
\vspace{-2mm}
\label{sec:experiments_and_results}
We conducted experiments on English and Spanish, using two synthesized datasets derived from the Conversational \cite{jia2019leveraging} and Common Voice 11 datasets \cite{ardila2020common} and well as the original Common Voice 11 datasets. See Tab.\ref{tbl:datasets}  for the details of each dataset. A comprehensive table of hyper-parameters, training details and number of parameters is available \textcolor{black}{on the project \textcolor{blue}{\href{ https://google-research.github.io/lingvo-lab/translatotron3}{website}}}. To reconstruct speech from the predicted spectrogram, we used WaveFit, a non-autoregressive neutral vocoder \cite{koizumi2022wavefit}. We used a cascaded S2ST system as a baseline. For its MT module, we used a nearest neighbor MT from \cite{lample2018word}.
\subsection{Metrics}
\textbf{BLEU score:} The translation performance was measured by BLEU on ASR transcriptions from the translated speech, compared to the corresponding reference translation text. Because ASR can make errors, such BLEU can be considered as a lower bound of the translation performance. \textbf{Mean Opinion Score (MOS):} To evaluate naturalness, we use the standard 5-point scale mean opinion score (MOS) in naturalness with human raters. When computing MOS, we only include ratings where headphones were used. We note that human raters also take into consideration the cohesion of the speech when rating utterances, thus the MOS scores are also affected by the quality of the translations. \textbf{Average cosine similarity of speaker embeddings (CS):} We use the speaker encoder of the PnG NAT TTS model \cite{morioka2022residual} to compute the speaker embedding of the input and of the translated output. Using cosine similarity we measure the distance between the embeddings and report the average over the whole test set.
\subsection{Synthesized Speech Data}
\textbf{Unpaired Conversational Dataset (UC):} We experimented using the proprietary dataset described in \cite{jia2019leveraging}. We synthesized both source and target speech using a state-of-the-art TTS model \cite{jia2021png,shen2020non} and vocoder \cite{kalchbrenner2018efficient}. To un-pair the dataset we split it into two halves, and use each of them separately as monolingual data. Tab.~\ref{table:bleu_mos_unsup} shows the experimental results. The proposed approach demonstrated substantial improvements over the baseline; $+13.27$ increase in BLEU for En$\rightarrow$Es and $+18.14$ increase in BLEU for Es$\rightarrow$En. With respect to the naturalness of the translated speech, the proposed approach outperformed the baseline with respect to MOS. When computing MOS, the human raters perceived the speech of the TTS baseline as intelligible in certain cases, and rated accordingly. Because of this, although the TTS system is state of the art, it received lower than state-of-the-art human ratings. Overall, the proposed approach outperformed the baseline in $
\sim$1 MOS score in Es$\rightarrow$En and in 0.22 points in En$\rightarrow$Es.
\textbf{Common Voice 11 (CV11):} We considered English and Spanish, and generated the data using the same TTS model and vocoder that was employed for the Conversational dataset. This was done to ensure consistency and eliminate any potential variations in the dataset that could impact the results of the study. For evaluation, we used the CVSS (CV) \cite{jia2022cvss} Spanish-English test-set.
The proposed approach demonstrated a notable improvement in performance when compared to the baseline. As illustrated in Tab.~\ref{table:bleu_mos_unsup}, the proposed approach achieved approximately $30$\% improvement in BLEU over the baseline. Specifically, in Spanish to English, the proposed approach exhibited an improvement of $+3.77$ in BLEU over the baseline, and in English to Spanish, it demonstrated an improvement of $+3.99$ in BLEU over the baseline. Furthermore, the proposed approach got an improvement of $+0.77$ MOS score over the baseline, in Es$\rightarrow$En and $+0.14$ MOS score in En$\rightarrow$Es.

\subsection{Real Speech Data (CVE)} For real speech experiment, we used Common Voice 11. Natural (non-synthesized) monolingual speech-text datasets both in English and Spanish were used for training. The evaluation of Spanish-English real speech Translation was conducted using real speech with verified translation from the CoVoST2 test set (CVE) \cite{wang2020covost}. It should be noted there is no English-Spanish CoVoST2 test set, therefore only an Spanish-English evaluation is presented. The proposed approach achieved an $10.67$ in BLEU for the task of Spanish-English Translation which is an improvement of  $+0.75$ in BLEU over the baseline which achieves $9.92$ BLEU. \textcolor{black}{Furthermore, the proposed approach got MOS of 4.21 points which is better then the baseline approach}. Audio samples are available in our \textbf{\href{https://google-research.github.io/lingvo-lab/translatotron3}{website}}.
\subsection{Comparison to Supervised Approaches} 
Tab.~\ref{table:bleu_mos_unsup} shows the performance of the cascade baseline, the proposed approach, and the supervised approaches on the CVSS dataset. For the English to Spanish translation task, the proposed approach demonstrated a level of performance that was comparable to a supervised approach. Notably, the margin by which we fell behind was relatively small, particularly when considering the fact that unsupervised translation is a more challenging task in comparison to supervised translation. On the other hand, for the Spanish to English translation task, the supervised approaches exhibited significantly better performance. While this margin may appear significant, it is worth noting that the difficulty of the task should be considered. Unlike Translatotron 2, which relies on parallel training pairs of aligned English and Spanish utterances, our proposed method avoids the need for explicitly annotated bilingual data. Each training instance within our approach comprises a single utterance, either Spanish or English.

\begin{table}[]
\centering
\begin{small}
\caption{Monolingual datasets used.}
\begin{tabular}{lccc}
    \toprule & UC \cite{jia2019leveraging} & CV \cite{ardila2020common} & CVE \cite{ardila2020common} \\
    \midrule
    Languages & Es $\leftrightarrow$ En & Es $\leftrightarrow$ En & Es $\leftrightarrow$ En \\
    Domain & Synthesized & Synthesized & Read,short-form \\
    SR [kHz] & 24 & 24 & 48 \\
    Spanish [h] & 350 & 340 & 413  \\
    English [h] & 371 & 427 & 708  \\
    Synthesizer & \multicolumn{2}{c}{PnGNAT $+$ WaveFit} & Real Speech \\

\bottomrule
\end{tabular}
\label{tbl:datasets}
\end{small}
\vspace{-3mm}
\end{table}

\subsection{Ablation Analysis} 
\vspace{-2mm}
An ablation study was conducted. This study involved the removal of specific components. 
(i) Removed the reconstruction loss Eq.~\eqref{eqn:recon_loss} (``-Recon''), and
(ii) removed Back-Translation loss (``-BT'') Eq.~\eqref{eq:back_translation}, and
(iii) removed the MUSE embeeding loss $\mathcal{L}_{\text{MUSE}}$ at the encoder output (``-Muse''), and (iv) disabled SpecAugment (``-SA'') \cite{park2019specaugment}. The results are summarized in Tab.~\ref{table:ablation}. It can be seen from the table that that each of the aforementioned components contributed to the overall performance improvement of the proposed approach, with the reconstruction loss and the back-translation loss having the most significant impact. Additionally, the inclusion of the MUSE loss in the model also resulted in a notable improvement in performance. 
\subsection{Para-/Non-Linguistic Feature Preservation} 
Tab.\ref{table:bleu_mos_unsup} reports the average similarity scores for the test set. It show that the proposed approach can preserve paralinguistic information during translation. The original speaker's voice is preserved to a high degree. For the baseline, we use a TTS system and pick speakers randomly. A value of ~$0.6$ shows that there is a good correlation between input and output speaker identities. The baseline values of ~$0.16$ shows that picking random speakers gives no correlation. For Real speech data, our approach also shows good correlation with a value of ~$0.34$. Moreover, our approach shows promise in maintaining critical elements such as pauses and speaking rates as can be examined in our \href{ https://google-research.github.io/lingvo-lab/translatotron3}{\textbf{website}}.

\begin{table}[]
\vspace{0mm}
\centering
\caption{Performance of Translatotron~3 on S2ST and comparison to supervised approaches.}
\label{table:bleu_mos_unsup}

\resizebox{\columnwidth}{!}{%
\begin{tabular}{llllllll}
\toprule
\multirow{1}{*}{DS} &
  \multirow{1}{*}{Model} &
  \multicolumn{3}{c}{En $\rightarrow$ Es} &
  \multicolumn{3}{c}{Es $\rightarrow$ En} \\ \cline{3-5} \cline{6-8} 
 &
  \multicolumn{1}{c}{} &
  \multicolumn{1}{r}{MOS} &
  \multicolumn{1}{r}{BLEU} &
  \multicolumn{1}{r}{CS} &
  \multicolumn{1}{r}{MOS} &
  \multicolumn{1}{r}{BLEU} &
  \multicolumn{1}{r}{CS} \\
\midrule
\multicolumn{3}{l}{\textbf{Unsupervised}}                      &     &      \\
\midrule
\multirow{2}{*}{UC} & Cascade & \multicolumn{1}{l}{3.48} & \multicolumn{1}{l}{5.58} & \multicolumn{1}{l}{0.15}  & \multicolumn{1}{l}{3.24} & \multicolumn{1}{l}{6.13}  & \multicolumn{1}{l}{0.16}  \\
                        & T3  & \multicolumn{1}{l}{3.70} &  \multicolumn{1}{l}{18.85} &  \multicolumn{1}{l}{0.62} & \multicolumn{1}{l}{4.20} & \multicolumn{1}{l}{24.27} & \multicolumn{1}{l}{0.65}  \\
\midrule
\multirow{2}{*}{CV11}  & Cascade & \multicolumn{1}{l}{3.64} & \multicolumn{1}{l}{9.46} &  \multicolumn{1}{l}{0.16}   & \multicolumn{1}{l}{3.36} & \multicolumn{1}{l}{10.48} & \multicolumn{1}{l}{0.17}  \\ 
                        & T3  & \multicolumn{1}{l}{3.78} & \multicolumn{1}{l}{13.45}  & \multicolumn{1}{l}{0.63} & \multicolumn{1}{l}{4.13} & \multicolumn{1}{l}{14.25} & \multicolumn{1}{l}{0.56}  \\
\midrule
\multirow{2}{*}{CVE}  & Cascade & \multicolumn{1}{l}{-} & \multicolumn{1}{l}{-} &  \multicolumn{1}{l}{-}   & \multicolumn{1}{l}{3.30} & \multicolumn{1}{l}{9.92} & \multicolumn{1}{l}{0.17}  \\ 
                        & T3  & \multicolumn{1}{l}{-} & \multicolumn{1}{l}{-}  & \multicolumn{1}{l}{-} & \multicolumn{1}{l}{4.21} & \multicolumn{1}{l}{10.76} & \multicolumn{1}{l}{0.34}  \\ 
\midrule
\multicolumn{3}{l}{\textbf{Supervised}}                      &     &      \\
\midrule
\multirow{7}{*}{CV11}  & T1\textsuperscript{\cite{jia2022cvss}}   & \multicolumn{1}{l}{-}&    \multicolumn{1}{l}{-}     & \multicolumn{1}{l}{-} & \multicolumn{1}{l}{-} &  \multicolumn{1}{l}{19.8} & \multicolumn{1}{l}{-}\\
                        & T2\textsuperscript{\cite{jia2022cvss}}    & \multicolumn{1}{l}{-}&    \multicolumn{1}{l}{-}    & \multicolumn{1}{l}{-} & \multicolumn{1}{l}{4.61} & \multicolumn{1}{l}{25.4} & \multicolumn{1}{l}{-}  \\

                        & TS\textsuperscript{\cite{huang2022transpeech}}                   & \multicolumn{1}{l}{-}&    \multicolumn{1}{l}{14.94}     & \multicolumn{1}{l}{-} & \multicolumn{1}{l}{-}  & \multicolumn{1}{l}{-}  & \multicolumn{1}{l}{-} \\
                        & S2TT\textsuperscript{\cite{inaguma2022unity}}        & \multicolumn{1}{l}{-}&    \multicolumn{1}{l}{-}    & \multicolumn{1}{l}{-}  & \multicolumn{1}{l}{-}  & \multicolumn{1}{l}{18.2} & \multicolumn{1}{l}{-}\\
                        & S2UT\textsuperscript{\cite{inaguma2022unity}}   & \multicolumn{1}{l}{-}&    \multicolumn{1}{l}{-}    & \multicolumn{1}{l}{-} & \multicolumn{1}{l}{-}  & \multicolumn{1}{l}{25.9} & \multicolumn{1}{l}{-}\\
                        & UnitY\textsuperscript{\cite{inaguma2022unity}}   & \multicolumn{1}{l}{-}&    \multicolumn{1}{l}{-}    & \multicolumn{1}{l}{-} & \multicolumn{1}{l}{-}  & \multicolumn{1}{l}{29.0} & \multicolumn{1}{l}{-}\\
\bottomrule

\end{tabular}}
\vspace{-5mm}
\end{table}

\begin{table}[]
\caption{Ablation analysis. BLEU scores.}
\label{table:ablation}
\resizebox{\columnwidth}{!}{%
\begin{tabular}{llllll}
\toprule
      & Full & -Recon & -BT & -Muse & -SA \\
\midrule
\textbf{En $\rightarrow$ Es} & 18.85    & 0.41   & 1.91       & 5.44      & 9.23     \\
\textbf{Es $\rightarrow$ En} & 24.27    & 2.99   & 3.62       & 6.22      & 12.88   \\
\bottomrule
\end{tabular}}
\vspace{0mm}
\end{table}

\section{Conclusion}
\vspace{-2mm}
\label{sec:conclusion}
Translatotron~3, is an unsupervised direct S2ST model. It uses the unsupervised embedding word mapping and a back-translation training procedure. Unlike the previous approaches, the proposed approach can implicitly preserve some elements of para-/non-linguistic characteristics in the source speech. Translatotron~3 improved upon the unsupervised cascade baseline (10.51 BLEU) and approached the performance of supervised systems on the CVSS dataset (by 1.95 gap in BLEU). This suggests that Translatotron~3 is an effective approach for unsupervised S2ST that is able to retain important information from the source speech in the translation.

\section{Acknowledgments}
\vspace{-2mm}
The authors would like to thank Nanxin Chen, Yuma Koizumi, Soroosh Mariooryad, RJ Skerry-Ryan, Neil Zeghidour, Christian Frank, Marco Tagliasacchi, Nadav Bar, and the rest of the Google Research team for helpful discussions and data preparation. The contribution of Alon Levkovitch is part of a Ph.D. thesis research conducted at Tel-Aviv University.

\clearpage
\bibliographystyle{IEEEbib}
\footnotesize
\bibliography{strings,refs}

\begin{thebibliography}{10}

\bibitem{jia2022translatotron}
Ye~Jia~et al,
\newblock ``Translatotron 2: High-quality direct speech-to-speech translation
  with voice preservation,''
\newblock in {\em Proc. ICML}, 2022, pp. 10120--10134.

\bibitem{lavie1997janus}
Alon Lavie~et al,
\newblock ``{JANUS-III}: Speech-to-speech translation in multiple languages,''
\newblock in {\em Proc. ICASSP}, 1997, vol.~1, pp. 99--102.

\bibitem{wahlster2013verbmobil}
Wolfgang Wahlster,
\newblock {\em Verbmobil: Foundations of speech-to-speech translation},
\newblock Springer Science \& Business Media, 2013.

\bibitem{nakamura2006atr}
Satoshi Nakamura~et al,
\newblock ``The {ATR} multilingual speech-to-speech translation system,''
\newblock {\em IEEE Trans. ASLP}, vol. 14, no. 2, pp. 365--376, 2006.

\bibitem{jia2019direct}
Ye~Jia~et al,
\newblock ``Direct speech-to-speech translation with a sequence-to-sequence
  model,''
\newblock {\em Proc. Interspeech}, pp. 1123--1127, 2019.

\bibitem{lee2021direct}
Ann Lee~et al,
\newblock ``Direct speech-to-speech translation with discrete units,''
\newblock {\em arXiv:2107.05604}, 2021.

\bibitem{lee-etal-2022-direct}
Ann Lee~et al,
\newblock ``Direct speech-to-speech translation with discrete units,''
\newblock in {\em Proc. ACL}, 2022, pp. 3327--3339.

\bibitem{sennrich2015improving}
Rico Sennrich~et al,
\newblock ``Improving neural machine translation models with monolingual
  data,''
\newblock {\em arXiv:1511.06709}, 2015.

\bibitem{artetxe2018unsupervised}
Mikel Artetxe~et al,
\newblock ``Unsupervised neural machine translation,''
\newblock in {\em Proc. ICLR}, 2018.

\bibitem{lample2018unsupervised}
Guillaume Lample~et al,
\newblock ``Unsupervised machine translation using monolingual corpora only,''
\newblock in {\em Proc. ICLR}, 2018.

\bibitem{he2022masked}
Kaiming He~et al,
\newblock ``Masked autoencoders are scalable vision learners,''
\newblock in {\em Proc. CVPR}, 2022, pp. 16000--16009.

\bibitem{park2019specaugment}
Daniel~S Park~et al,
\newblock ``Specaugment: A simple data augmentation method for automatic speech
  recognition,''
\newblock {\em Proc. Interspeech}, pp. 2613--2617, 2019.

\bibitem{conneau2017word}
Alexis Conneau~et al,
\newblock ``Word translation without parallel data,''
\newblock {\em arXiv:1710.04087}, 2017.

\bibitem{lample2017unsupervised}
Guillaume Lample, Alexis Conneau, Ludovic Denoyer, and Marc'Aurelio Ranzato,
\newblock ``Unsupervised machine translation using monolingual corpora only,''
\newblock {\em arXiv preprint arXiv:1711.00043}, 2017.

\bibitem{zhang2021uwspeech}
Chen Zhang~et al,
\newblock ``Uwspeech: Speech to speech translation for unwritten languages,''
\newblock in {\em Proc. AAAI}, 2021, vol.~35, pp. 14319--14327.

\bibitem{lee2021textless}
Ann Lee~et al,
\newblock ``Textless speech-to-speech translation on real data,''
\newblock {\em arXiv:2112.08352}, 2021.

\bibitem{huang2022transpeech}
Rongjie Huang~et al,
\newblock ``{TranSpeech}: Speech-to-speech translation with bilateral
  perturbation,''
\newblock {\em arXiv:2205.12523}, 2022.

\bibitem{tang2022unified}
Yun Tang~et al,
\newblock ``Unified speech-text pre-training for speech translation and
  recognition,''
\newblock in {\em Proc. ACL}, 2022, pp. 1488--1499.

\bibitem{dong2022leveraging}
Qianqian Dong~et al,
\newblock ``Leveraging pseudo-labeled data to improve direct speech-to-speech
  translation,''
\newblock {\em arXiv:2205.08993}, 2022.

\bibitem{kano2021transformer}
Takatomo Kano~et al,
\newblock ``Transformer-based direct speech-to-speech translation with
  transcoder,''
\newblock in {\em Proc. IEEE SLT}, 2021, pp. 958--965.

\bibitem{wang2022simple}
Changhan Wang~et al,
\newblock ``Simple and effective unsupervised speech translation,''
\newblock {\em arXiv:2210.10191}, 2022.

\bibitem{inaguma2022unity}
Hirofumi Inaguma~et al,
\newblock ``Unity: Two-pass direct speech-to-speech translation with discrete
  units,''
\newblock {\em arXiv preprint arXiv:2212.08055}, 2022.

\bibitem{wei2022joint}
Kun Wei~et al,
\newblock ``Joint pre-training with speech and bilingual text for direct speech
  to speech translation,''
\newblock {\em arXiv:2210.17027}, 2022.

\bibitem{zeghidour22-soundstream}
Neil Zeghidour, Alejandro Luebs, Ahmed Omran, Jan Skoglund, and Marco
  Tagliasacchi,
\newblock ``Soundstream: An end-to-end neural audio codec,''
\newblock {\em {IEEE} {ACM} Trans. Audio Speech Lang. Process.}, vol. 30, pp.
  495--507, 2022.

\bibitem{chung2021w2v}
Yu-An Chung~et al,
\newblock ``W2v-bert: Combining contrastive learning and masked language
  modeling for self-supervised speech pre-training,''
\newblock in {\em 2021 IEEE Automatic Speech Recognition and Understanding
  Workshop (ASRU)}. IEEE, 2021, pp. 244--250.

\bibitem{defossez2022high}
Alexandre D{\'e}fossez~et al,
\newblock ``High fidelity neural audio compression,''
\newblock {\em arXiv preprint arXiv:2210.13438}, 2022.

\bibitem{baevski2020wav2vec}
Alexei Baevski~et al,
\newblock ``wav2vec 2.0: A framework for self-supervised learning of speech
  representations,''
\newblock {\em Advances in neural information processing systems}, vol. 33, pp.
  12449--12460, 2020.

\bibitem{hsu2021hubert}
Wei-Ning Hsu~et al,
\newblock ``Hubert: Self-supervised speech representation learning by masked
  prediction of hidden units,''
\newblock {\em IEEE/ACM Transactions on Audio, Speech, and Language
  Processing}, vol. 29, pp. 3451--3460, 2021.

\bibitem{tjandra2019speech}
Andros Tjandra, Sakriani Sakti, and Satoshi Nakamura,
\newblock ``Speech-to-speech translation between untranscribed unknown
  languages,''
\newblock in {\em 2019 IEEE Automatic Speech Recognition and Understanding
  Workshop (ASRU)}. IEEE, 2019, pp. 593--600.

\bibitem{rubenstein2023audiopalm}
Paul~K Rubenstein~et al,
\newblock ``Audiopalm: A large language model that can speak and listen,''
\newblock {\em arXiv preprint arXiv:2306.12925}, 2023.

\bibitem{li2023textless}
Xinjian Li~et al,
\newblock ``Textless direct speech-to-speech translation with discrete speech
  representation,''
\newblock in {\em ICASSP 2023-2023 IEEE International Conference on Acoustics,
  Speech and Signal Processing (ICASSP)}. IEEE, 2023, pp. 1--5.

\bibitem{kim2023many}
Minsu Kim~et al,
\newblock ``Many-to-many spoken language translation via unified speech and
  text representation learning with unit-to-unit translation,''
\newblock {\em arXiv preprint arXiv:2308.01831}, 2023.

\bibitem{wei2023joint}
Kun Wei~et al,
\newblock ``Joint pre-training with speech and bilingual text for direct speech
  to speech translation,''
\newblock in {\em ICASSP 2023-2023 IEEE International Conference on Acoustics,
  Speech and Signal Processing (ICASSP)}. IEEE, 2023, pp. 1--5.

\bibitem{zhang2023dub}
Dong Zhang~et al,
\newblock ``Dub: Discrete unit back-translation for speech translation,''
\newblock {\em arXiv preprint arXiv:2305.11411}, 2023.

\bibitem{ao2021speecht5}
Junyi Ao~et al,
\newblock ``Speecht5: Unified-modal encoder-decoder pre-training for spoken
  language processing,''
\newblock {\em arXiv preprint arXiv:2110.07205}, 2021.

\bibitem{zhang2022yitrans}
Ziqiang Zhang~et al,
\newblock ``The yitrans end-to-end speech translation system for iwslt 2022
  offline shared task,''
\newblock {\em arXiv preprint arXiv:2206.05777}, 2022.

\bibitem{zhang2022speechlm}
Ziqiang Zhang~et al,
\newblock ``Speechlm: Enhanced speech pre-training with unpaired textual
  data,''
\newblock {\em arXiv preprint arXiv:2209.15329}, 2022.

\bibitem{artetxe2017learning}
Mikel Artetxe~et al,
\newblock ``Learning bilingual word embeddings with (almost) no bilingual
  data,''
\newblock in {\em Proc. ACL}, 2017, pp. 451--462.

\bibitem{artetxe2018robust}
Mikel Artetxe~et al,
\newblock ``A robust self-learning method for fully unsupervised cross-lingual
  mappings of word embeddings,''
\newblock in {\em Proc. ACL}, 2018.

\bibitem{chung2019towards}
Yu-An Chung, Wei-Hung Weng, Schrasing Tong, and James Glass,
\newblock ``Towards unsupervised speech-to-text translation,''
\newblock in {\em ICASSP 2019-2019 IEEE International Conference on Acoustics,
  Speech and Signal Processing (ICASSP)}. IEEE, 2019, pp. 7170--7174.

\bibitem{zhu2017unpaired}
Jun-Yan Zhu~et al,
\newblock ``Unpaired image-to-image translation using cycle-consistent
  adversarial networks,''
\newblock in {\em Proc. ICCV}, 2017, pp. 2223--2232.

\bibitem{jia2019leveraging}
Ye~Jia~et al,
\newblock ``Leveraging weakly supervised data to improve end-to-end
  speech-to-text translation,''
\newblock in {\em Proc. ICASSP}, 2019, pp. 7180--7184.

\bibitem{ardila2020common}
Rosana Ardila~et al,
\newblock ``{Common Voice}: A massively-multilingual speech corpus,''
\newblock in {\em Proc. LREC}, 2020.

\bibitem{koizumi2022wavefit}
Yuma Koizumi~et al,
\newblock ``{WaveFit}: An iterative and non-autoregressive neural vocoder based
  on fixed-point iteration,''
\newblock {\em arXiv:2210.01029}, 2022.

\bibitem{lample2018word}
Guillaume Lample~et al,
\newblock ``Word translation without parallel data,''
\newblock in {\em Proc. ICLR}, 2018.

\bibitem{morioka2022residual}
Nobuyuki Morioka~et al,
\newblock ``Residual adapters for few-shot text-to-speech speaker adaptation,''
\newblock {\em arXiv preprint arXiv:2210.15868}, 2022.

\bibitem{jia2021png}
Ye~Jia~et al,
\newblock ``{PnG BERT}: Augmented {BERT} on phonemes and graphemes for neural
  {TTS},''
\newblock in {\em Proc. Interspeech}, 2021.

\bibitem{shen2020non}
Jonathan Shen~et al,
\newblock ``{Non-Attentive Tacotron}: Robust and controllable neural {TTS}
  synthesis including unsupervised duration modeling,''
\newblock {\em arXiv:2010.04301}, 2020.

\bibitem{kalchbrenner2018efficient}
N.~Kalchbrenner~et al,
\newblock ``Efficient neural audio synthesis,''
\newblock in {\em Proc. ICML}, 2018.

\bibitem{jia2022cvss}
Ye~Jia~et al,
\newblock ``{CVSS} corpus and massively multilingual speech-to-speech
  translation,''
\newblock {\em arXiv:2201.03713}, 2022.

\bibitem{wang2020covost}
Changhan Wang~et al,
\newblock ``{CoVoST} 2 and massively multilingual speech-to-text translation,''
\newblock {\em arXiv:2007.10310}, 2020.

\end{thebibliography}

\end{document}